\documentclass[10pt,twocolumn,letterpaper]{article}

\usepackage[final]{cvpr}      
\usepackage{graphicx}
\usepackage{subcaption} 
\usepackage{caption} 
\usepackage[accsupp]{axessibility}
\usepackage[normalem]{ulem}
\usepackage{xcolor}

\usepackage{array}
\newcolumntype{x}[1]{>{\centering\arraybackslash}p{#1}}
\newcolumntype{y}[1]{>{\raggedright\arraybackslash}p{#1}}
\newcolumntype{z}[1]{>{\raggedleft\arraybackslash}p{#1}}
\newcommand{\tablestyle}[2]{\setlength{\tabcolsep}{#1}\renewcommand{\arraystretch}{#2}\centering\footnotesize}

\newcommand{\myparagraph}[1]{\noindent\textbf{#1}}

\definecolor{cvprblue}{rgb}{0.21,0.49,0.74}
\usepackage[pagebackref,breaklinks,colorlinks,allcolors=cvprblue]{hyperref}

\def\paperID{20} 
\def\confName{CVPR}
\def\confYear{2025}

\title{SmallGS: Gaussian Splatting-based Camera Pose Estimation \\for Small-Baseline Videos }

\author{
  Yuxin Yao$^{1}$,\;  Yan Zhang$^{2}$,\; Zhening Huang$^1$,\; Joan Lasenby$^{1}$ \\
  $^1$University of Cambridge,\;
  $^2$Meshcapade \\ 
  {\tt\small \{yy561,zh340,jl221\}@cam.ac.uk}; {\tt\small yan@meshcapade.com}
}

\begin{document}

\twocolumn[{%
\renewcommand\twocolumn[1][]{#1}%
\maketitle
\begin{center}
    \captionsetup{type=figure}
    \includegraphics[width=\linewidth]{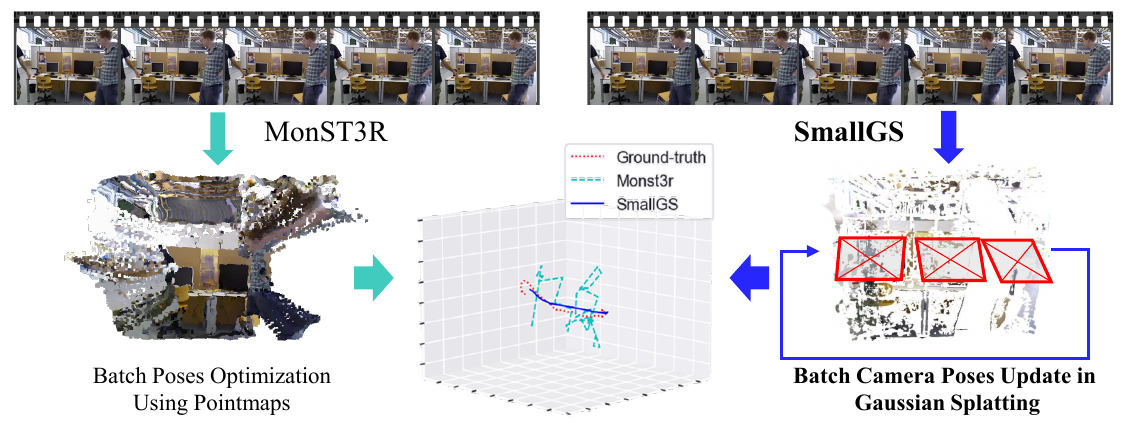}
    \captionof{figure}{\textbf{Camera pose estimation for small-baseline videos with SmallGS.} Our method focuses on camera pose estimation in small-baseline videos, updating the camera poses based on the rasterization of Gaussian splatting in the learned camera viewpoints. We achieved better and smoother results compared to the previous SOTA MonST3R. 
    }\label{fig:teaser}
\end{center}%
}]


\begin{abstract}
Dynamic videos with small baseline motions are ubiquitous in daily life, especially on social media. 
However, these videos present a challenge to existing pose estimation frameworks due to ambiguous features, drift accumulation, and insufficient triangulation constraints.
Gaussian splatting, which maintains an explicit representation for scenes, provides a reliable novel view rasterization when the viewpoint change is small.
Inspired by this, we propose \textbf{SmallGS}, a camera pose estimation framework that is specifically designed for small-baseline videos. SmallGS optimizes sequential camera poses using Gaussian splatting, which reconstructs the scene from the first frame in each video segment to provide a stable reference for the rest. The temporal consistency of Gaussian splatting within limited viewpoint differences reduced the requirement of sufficient depth variations in traditional camera pose estimation. 
We further incorporate pretrained robust visual features, e.g. DINOv2, into Gaussian splatting, where high-dimensional feature map rendering enhances the robustness of camera pose estimation. By freezing the Gaussian splatting and optimizing camera viewpoints based on rasterized features, SmallGS effectively learns camera poses without requiring explicit feature correspondences or strong parallax motion.
We verify the effectiveness of SmallGS in small-baseline videos in TUM-Dynamics sequences, which achieves impressive accuracy in camera pose estimation compared to MonST3R and DORID-SLAM for small-baseline videos in dynamic scenes. Our project page is at: \href{https://yuxinyao620.github.io/SmallGS}{https://yuxinyao620.github.io/SmallGS}

\end{abstract}

 \section{Introduction}
\label{sec:intro}
With the advancement of deep learning and computer vision, camera pose estimation has improved significantly over the past decade. Structure-from-Motion (SfM) and SLAM-based methods represent mature approaches for estimating camera poses in static scenes with large camera baselines \cite{megasam}. However, videos captured in everyday scenarios are typically recorded using stationary setups, exhibiting small baselines and limited viewpoint rotation. These characteristics result in substantial camera drift accumulation 
as trajectories extend. Although bundle adjustment requires sufficient geometric constraints to mitigate drift, videos with small baselines inherently lack diverse camera motion, which limits optimization effectiveness.

Furthermore, recent methods that aim to reconstruct global 3D point clouds thereby capture the entire scene’s geometry for better camera pose estimation. Deep visual SLAM methods \cite{monst3r,dust3r} leverage visual features for robust camera tracking and scene reconstruction. However, most struggle with videos featuring small-baselines and unconstrained camera trajectories \cite{schoenberger2016sfm,schoenberger2016mvs}.

Gaussian splatting as an explicit scene representation provides a feasible way to update camera poses while reconstructing the 3D scene.
Compared to NeRF-based methods, Gaussian splatting offers a more efficient way to optimize camera pose, since affine transformations can be directly applied to 3D Gaussians. Camera pose estimation via 3D Gaussian splatting is less susceptible to small-baseline limitations, as it optimizes relative poses between adjacent frames. However, most methods \cite{cf3dgs,instantsplat,zerogs} focus exclusively on static scenes, and their iterative pairwise camera pose estimation remains computationally expensive.

In this work, we focus on camera pose estimation in small-baseline videos, where the camera motion exhibits limited translations. Such videos are ubiquitous on social medias like TikTok and Instagram.
Camera poses and intrinsic parameters can be estimated by leveraging advanced global visual features and bundle adjustment \cite{monst3r,dust3r}. While these approaches can produce realiable
global scene geometry and camera intrinsics, the estimated poses exhibit low precision with significant jitter. 


We introduce SmallGS for accurate camera tracking in in-the-wild dynamic videos with small camera movements, and leverage DINOv2 \cite{dinov2} features in the 3D Gaussians in order to improve robustness. 
Following CF-3DGS \cite{cf3dgs}, we lift depth maps predicted by MonST3R to 3D dense point clouds to initialize Gaussian splatting.  
To address moving objects in the scene, we mask dynamic regions using MonST3R's predicted confidence masks, which indicate the likelihood of an object being in motion. 

Although MonST3R can also estimate camera poses, the results in our small-baseline scenarios are not sufficiently accurate and have artifacts like jittering.
MonST3R estimates camera poses primarily through pointmap alignment, where pointmaps are generated by a neural network. However, in small-baseline videos, key constraints—such as sufficient depth variation—are often insufficient, making it difficult for the neural network to produce consistent and robust pointmaps. 
This will lead to e.g. inaccurate estimates of the pointmaps that are far away from the camera, and hence will negatively influence the camera poses that are obtained via pointmap alignment, illustrated in Fig. \ref{fig:teaser}

In contrast, Gaussian splatting leverages neural rendering and photometric losses to estimate the camera SE(3) transformations \cite{cf3dgs}. The splatting process will focus on the 3D Gaussians that are closer to the camera and have higher opacities. In this case, inaccuracies of scene geometries that are far away from the camera will not have significant impact. Small baselines that are with respect to the far objects are actually \textit{not small} with respect to objects in the near field.
In addition, since adjacent frames exhibit high similarity, frames can be represented using the Gaussian splatting reconstructed from the neighboring frame. To further exploit this consistency, we employ a sliding window to partition the image sequences into segments, where the first frame of each segment is the same as the last frame of the previous segment. 
The Gaussian splatting for the first frame is optimized to match the RGB image, reducing the reliance on depth predictions within the batch and improving pose stability(Fig. \ref{fig:teaser}).
We then jointly optimize a batch of adjacent camera poses using a single Gaussian splatting scene, significantly improving computational efficiency. 
Furthermore, we incorporate a camera trajectory smoothness constraint to encourage smooth camera motion and remove jittering artifacts.


We evaluate our method on the TUM-dynamics dataset \cite{sturm2012benchmark}, which consists of handheld camera recordings in dynamic environments. SmallGS estimates camera poses using MonST3R’s predicted depth maps, semantic masks, camera intrinsics, and DINOv2 features. Our approach significantly improves Absolute Trajectory Error ($ATE$) and Relative Pose Error ($RPE$). Additionally, we analyze the velocity difference between the estimated and ground-truth camera positions, providing insights into the smoothness of the estimated camera trajectory. The velocity difference also serves as a measure of the motion consistency between the estimated and ground-truth camera trajectories.

In summary, the main contributions of our work are as follows.
\begin{itemize}
    \item \textbf{Establish a novel camera pose estimation framework, SmallGS, for small-baseline videos}, leveraging Gaussian splatting to optimize camera poses while mitigating dynamic object interference via predicted semantic masks. SmallGS does not rely on 3D alignments or triangulation, alleviating the instability in camera pose estimation caused by limited parallax and weak geometric constraints.
    \item \textbf{Utilize visual features from DINOv2 in camera pose estimation}, investigating the influence of geometry understanding and higher-dimensional features. This improves the accuracy of camera pose estimation.

\end{itemize}

\section{Related Works}
\subsection{3D Gaussian splatting}
3D Gaussian splatting is a novel scene representation method, which is explicit, fast, and capable of high-quality rendering. A set of 3D Gaussians with different covarainces and means are used to represent static scenes \cite{gs,cf3dgs,featuregs,Scaffold-GS,instantsplat}. By incorporating time-varying parameters in each Gaussian, it can also effectively represent dynamic scenes \cite{luiten2023dynamic, Wu_2024_CVPR, SplatFields}. Sparse points and camera poses for each image are required for scene reconstruction, which is commonly obtained using COLMAP \cite{schoenberger2016sfm, schoenberger2016mvs}. However, COLMAP relies on matching the extracted feature descriptor and triangulation to reconstruct 3D scenes \cite{Scaffold-GS, instantsplat}, which can lead to inaccuracies when estimating depth and camera pose in videos with small baseline disparity.

To resolve preprocessing requirements, CF-3DGS \cite{cf3dgs} utilized pre-trained depth estimation models like DPT \cite{dpt} to lift image pixels into a dense point cloud. CF-3DGS progressively grows the 3D Gaussian set by processing one neighbouring frame at a time and estimating camera poses through optimization without relying on COLMAP. Similarly, InstantSplat \cite{instantsplat} obtains relative camera poses by aligning dense pointmaps predicted using DUSt3r \cite{instantsplat}.

CF-3DGS and InstantSplat are primarily applied to static scenes. Dynamic 3DGS enables dynamic reconstruction by applying the dense tracking of Gaussians \cite{luiten2023dynamic}. SplatFields employ DOMA \cite{DOMA} to learn the motion fields of the Gaussians, introducing spatial bias which stabilizes the optimization process.
However, most dynamic scene methods require pre-established camera motion models to estimate camera poses incrementally \cite{luiten2023dynamic}.

\subsection{Traditional Camera Pose Estimation Methods}
Estimating camera pose is one of the most critical components in the pipelines of 3D reconstruction, robotics, and AR/VR systems. The estimation of camera extrinsic and intrinsic parameters involves techniques like SfM and SLAM, primarily used for cases of unstructured video sequences. Traditional SLAM methods rely on aligning image correspondences or utilizing feature descriptors to establish alignment between different viewpoints, followed by bundle adjustment to optimize 3D point cloud positions and camera parameters \cite{davison2007monoslam,mourikis2007multi,engel2014lsd,newcombe2011dtam}. Recent advances in deep learning have enabled SLAM and SfM approaches to learn higher-dimensional features for matching and scene reconstruction, which enhances robustness to blur and artifacts while reducing the need for extensive preprocessing \cite{teed2021droid,detone2018superpoint,sucar2021imap}. DROID-SLAM \cite{teed2021droid} introduces a recurrent bundle adjustment layer using GRU, effectively integrating camera pose estimation with pixel-wise depth recovery. However, these methods mainly focus on static scenes and require large baselines to establish sufficient triangulation constraints. Consequently, their performance degrades in dynamic environments or scenarios with small baseline video sequences, which are prevalent in everyday applications.

To address these problems, SLAM and SfM methods incoorperated pretrained depth and semantic segmentation neural networks. The depth map, semantic segmentation, and camera poses are optimized jointly, improving the overall geometry of the scene and exploiting the epipolar constriant with respect to the static parts of the scene \cite{flowfusion,tan2013robust,casualslam,kopf2021robust}. Casual-SLAM \cite{casualslam} jointly optimized the depth and camera pose estimation. It fine-tuned a pre-trained depth network to represent dense 3D correspondance. Robust-CVD \cite{kopf2021robust} also jointly estimate camera parameters and depth maps from dynamic videos, while the optimization is based on the alignment of spatially-varying splines.  
\subsection{Learning-based Camera Pose Estimation}
Predicting the camera parameters directly from an end-to-end deep learning architecture is increasingly powerful. Learning-based visual odometry methods allow large-scale training under different scenes, static and dynamic \cite{shen2023dytanvo,chen2024leap,mur2015orb,mur2017orb,dust3r,monst3r}. For example, DytanVO \cite{shen2023dytanvo} developed a unified neural network that estimates the motion of the camera and segments dynamic objects, which is robust in dynamic videos. 
Similarly, LEAP-VO \cite{chen2024leap} leverages the attention mechanism to learn long-term point tracking, inferring moving object masks for accurate camera pose estimation. 
However, the camera pose estimation of dynamic video with small baselines remains a problem. Drift accumulation and loop closure issues occur in learning-based camera parameter predictions. 

DUSt3r \cite{dust3r} directly regresses the geometry of the scene by learning the pointmap representation, which estimates the 3D location of each associated pixel in a pair of images in a common coordinate shared by the pair of frames. Downstream applications like camera pose estimation and 3D reconstruction can be achieved using the pointmaps. MonST3R \cite{monst3r}, fine-tuned from DUSt3r, achieves SOTA results of camera parameter estimation and 4D reconstruction in dynamic videos. The drift accumulation flaw is less serious in MonST3R. 

The concurrent work, GS-CPR \cite{liu2025gscpr}, utilized the fast rendering property of Gaussian splatting to refine camera poses. However, they pre-train a Gaussian splatting, followed by finetuning the prior camera pose based on the 2D-2D matching between query image and rendered image. Another concurrent work, ZeroGS \cite{zerogs}, focuses on static scene reconstruction with unposed and unordered images. Both of them ignored the small-baseline video cases.

Our method further improves the camera pose estimation performances in small baseline videos by utilizing  Gaussian splatting and DINOv2 visual features. We apply a method similar to CF-3DGS given the higher-dimensional visual features, leading to a more accurate and smooth camera pose in small baseline videos. 
\section{Method}
\begin{figure*}
    \centering
    \includegraphics[width=0.88\linewidth]{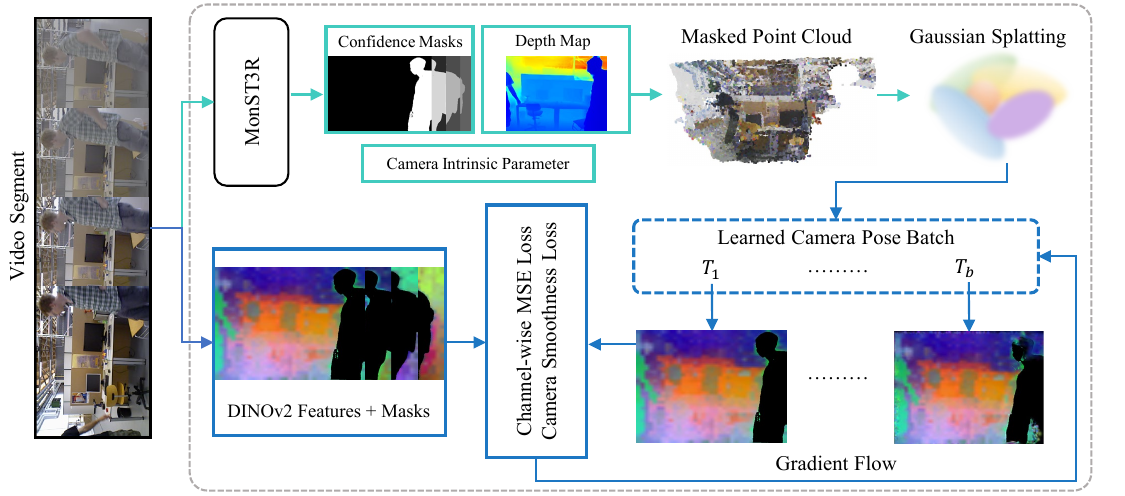}
    \caption{\textbf{Pipeline of SmallGS}. Our method follows the CF-3DGS pipeline, estimating camera poses in video segments. The process is: (1) Use MonST3R to predict depth maps, confidence masks, and camera intrinsics; (2) Lift the first frame's depth map into a dense point cloud, masking dynamic objects using the confidence mask as a semantic mask; (3) Initialize and update Gaussian splatting with the first frame; (4) Freeze the Gaussian parameters and optimize batched camera poses by minimizing the error between the rasterized feature maps (under the estimated poses) and the DINOv2~\cite{dinov2} feature maps, with semantic masks applied to both.}

    \label{fig:enter-label}
\end{figure*}
Given a sequence of unposed images in a dynamic video with a small camera baseline, we aim to estimate the camera poses. MonST3R provides robust pairwise pointmap predictions, along with downstream intrinsic and relative camera pose estimation, depth estimation, etc. Building upon existing methods, we propose SmallGS to learn the relative camera pose. In this section, we detail our approach. Sect.\ref{sec:method:pre} reviews the rendering process and camera optimization process of CF-3DGS \cite{cf3dgs}, while Sect.~\ref{sec:method:main} introduces SmallGS, which optimizes camera poses in batches to improve efficiency. Additionally, we incorporate a pre-trained visual feature extractor, DINOv2 \cite{dinov2}, into the SmallGS pipeline. The entire pipeline is shown in Fig. \ref{fig:enter-label}.
\subsection{Preliminary Colmap-Free 3D Gaussian splatting}
\label{sec:method:pre}
3D Gaussian splatting \cite{gs} represents scenes using 3D Gaussians, each characterized by several parameters: mean point ($\mu \in \mathbb{R}^3$), rotation factor ($r\in \mathbb{R}^4$), scale factor ($s\in \mathbb{R}^3$), opacity ($\alpha \in \mathbb{R}$), and spherical harmonics (SH) coefficients ($c\in \mathbb{R}^k$) representing color or features with $k$ degrees of freedom. The covariance matrix $\Sigma$ of a 3D Gaussian describes the corresponding 3D ellipsoid, which is determined by the scale factor and the rotation factor represented as a quaternion.
{\setlength{\abovedisplayskip}{0pt}
 \setlength{\belowdisplayskip}{0pt}
\begin{equation}
    \Sigma = RSS^TR^T
\end{equation}
}
where $R$ is the rotation matrix converted from the quaternion $r$, and $S$ is the diagonal matrix with the scale factor $s$, with $S = diag([s_x, s_y, s_z])$.

In CF-3DGS, the parameters and camera poses are updated by applying supervision on the rendered images. The rendering process of Gaussian splatting is differentiable \cite{gs} with respect to the camera pose $W \in \mathbf{SE(3)}$. The covariance of the Gaussians is projected onto the camera coordinates, resulting in 2D images at the camera viewpoint $W$ \cite{gs}.
{\setlength{\abovedisplayskip}{-1pt}
 \setlength{\belowdisplayskip}{-1pt}
\begin{equation}
    \Sigma_{2D} = JW\Sigma W^TJ^T 
\end{equation}
}
$J$ is the Jacobian of the affine approximation of the projective transformation. The color and opacity of each pixel in the rendered image are calculated by the alpha-blending of $N$ ordered points that overlap the pixel  \cite{gs}:
{\setlength{\abovedisplayskip}{0pt}
 \setlength{\belowdisplayskip}{0pt}
\begin{equation}
    C = \sum ^{N}_{i}c_i\alpha_i \prod^{i-1}_{j}(1-\alpha_j)
    \label{eq:color}
\end{equation}
}
where $C$ is the color for the pixel in rendered image. 
Similarly, rendering the higher-dimensional feature map $F$ instead of an image, Eq.\ref{eq:color} also applies  \cite{featuregs,gsplat}. 

The camera pose estimation of the images is achieved by optimizing the local relative camera poses of the adjacent frames. 3DGS provides an explicit representation of the entire scene, and the projection of each Gaussian in the scene relies on the viewing transform $W$, which is the camera pose \cite{cf3dgs}. Thus, the 2D projection of the same Gaussian under a different camera pose $W'$ is given by:
{\setlength{\abovedisplayskip}{2pt}
 \setlength{\belowdisplayskip}{2pt}
\begin{align}
    \mu_{2D} =  K\frac{W\mu}{(W\mu)_z} , \quad
    \mu'_{2D}= K \frac{W'\mu}{(W'\mu)_z}
\end{align}
}
where $(W\mu)_z$ denotes the depth (z-axis) of the Gaussian. $\mu'_{2D}$ denotes the pixel position of the projected Gaussians from a different viewpoint. $K$ denotes the camera intrinsic matrix. Hence, the camera pose can be optimized similarly to optimizing the center points of Gaussians  \cite{cf3dgs}. 

In CF-3DGS, the sparse points used for initialization are replaced by lifting the predicted depth using an off-the-shelf monocular depth estimation network, DPT  \cite{dpt}. With dense depth information, the geometry of Gaussian splatting closely approximates the target scene before optimization. The 3D Gaussians $G_t$ are further updated by minimizing the photometric loss between the rendered image and the current frame $t$. 
After Gaussian splatting is initialized with the first frame of the video, the relative camera transformation between frame $t$ and frame $t+1$ is optimized instead of the parameters of $G_t$.
{\setlength{\abovedisplayskip}{2pt}
 \setlength{\belowdisplayskip}{2pt}
\begin{equation}
T_t^* = \arg\min_{T_t} L_{rgb} (\mathcal{R}(T_t \odot G_t), I_{t+1})
\end{equation}
}
where $T_t$ is the relative camera transformation, including the rotation matrix and the translation vector, and $G_{t+1} = T_t \odot G_t$. The $\odot$ denotes the element-wise transformation. Since the two adjacent frames are close, the Gaussian splatting maintains robust visual features for camera pose optimization.

\subsection{Batch camera pose optimization}
To increase the efficiency of camera pose estimation in CF-3DGS for small-baseline videos, we introduce SmallGS, which optimizes batched camera poses. Small-baseline videos typically exhibit minimal camera movement over a few seconds, resulting in frames that remain highly similar to subsequent frames. A sliding window with size of $b$ is used to divide the video into video segments, where the first frame of each segment is the same as the last frame of the previous segment. Gaussian splatting is reinitialized for each video segment, and the relative camera poses are estimated.
{\setlength{\abovedisplayskip}{1pt}
 \setlength{\belowdisplayskip}{2pt}
\begin{equation}
    T_{t,..., t+b}^* = \arg\min_{T_t, ..., T_{t+b}} \sum_{i = t}^{t+b}L_{rgb} (\mathcal{R}(T_i \odot G_t), I_{i})
\end{equation}
}
where the camera intrinsics in the rendering function are given by MonST3R. The loss function $L_{rgb}$ is the mean squared error (MSE) loss between the rendered image and the ground-truth video frame. Upon estimating the relative camera poses for the entire sliding window, the relative camera pose for the full trajectory is obtained. However, in the presence of dynamic objects in the video, frame-to-frame consistency is broken. To address this, we introduce semantic masks to remove dynamic objects. MonST3R provides a confidence map for static objects in each image, and can serve as a semantic mask $M$ that masks out the dynamic objects. Additionally, we utilize the depth map predicted by MonST3R to initialize the point cloud for Gaussian splatting. Thus, each Gaussian splatting instance is initialized with both a depth map and a semantic mask. The image pixels, $p_{2D}$, of the first frame in the video segment are lifted to 3D camera coordinates using the corresponding depth map $D$. Then, the semantic mask $M$ is applied to filter out dynamic points.
{\setlength{\abovedisplayskip}{1pt}
 \setlength{\belowdisplayskip}{2pt}
\begin{equation}
    \mathbf{p} = \Pi^{-1} \left( K^{-1} \tilde{\mathbf{p}}_{2D} \cdot D \right) \cdot M
\end{equation}
}
where $\Pi^{-1}$ denotes the unprojection function that maps image coordinates and depth to 3D space, and $\tilde{\mathbf{p}}_{2D}$ represents the homogeneous image coordinates.

However, the batched camera poses within a video segment may have different scales, and the poses at the end of the window tend to exhibit more pronounced rotation and translation variations. To address this, a camera smoothness loss is applied during the batched optimization process to penalize large changes in camera pose. 
{\setlength{\abovedisplayskip}{1pt}
 \setlength{\belowdisplayskip}{1pt}
\begin{equation}
    L_{\text{smooth}} = \lambda_c\sum_{i=1}^{N-2} \left\| (\mathbf{x}_{i+1} - \mathbf{x}_{i}) - (\mathbf{x}_{i} - \mathbf{x}_{i-1}) \right\|
\end{equation}
}
where $\lambda_c$ varies from 0 to 1, increasing with the number of Gaussian splatting optimization iterations. $\mathbf{x}$ denotes the position of the camera. In addition, we use SSIM loss to measure the quality of the rendered image, with $\lambda_s$ set to 0.2 in our experiments.  
Overall, the optimization of batched camera poses is:
{\setlength{\abovedisplayskip}{-2pt}
 \setlength{\belowdisplayskip}{0pt}
\begin{multline}
    T_{t}^*...T^*_{t+b} = \arg\min_{T_t,..., T_{t+b}} \sum_{i = t}^{t+b}L_{rgb} (\mathcal{R}(T_i \odot G_t), I_{i})\\ 
    + L_{smooth} + \lambda_s L_{ssim}
\end{multline}
}

Using a robust visual feature instead of RGB images for Gaussian splatting enhances the overall scene geometry and semantics understanding \cite{featuregs}. Visual features are less affected by lighting or perspective changes. We utilize DINOv2  \cite{dinov2} features, which are generally robust in daily scenes, and select the $f$ most informative feature channels. Principal Component Analysis (PCA) is applied to rank the importance of features in each channel. The rendering process follows a similar approach to rendering color but with higher-dimensional features \cite{gsplat}. Thus, the optimization of batched camera poses is:
{\setlength{\abovedisplayskip}{-2pt}
 \setlength{\belowdisplayskip}{0pt}
\begin{multline}
    T_{t}^*...T^*_{t+b} = \arg\min_{T_t,..., T_{t+b}} \sum_{i = t}^{t+b}L_{mse} (\mathcal{R}(T_i \odot G_t), F^f_{i})\\  + L_{smooth} + \lambda_s L_{ssim}
\end{multline}
}
where $F^{f}_{i}$ represents the DINOv2 feature with $f$ selected channels. The loss function $L_{mse}$ in the optimization process is the channel-wise mean squared error (MSE) loss.

\label{sec:method:main}

\section{Experiment}
\begin{table*}[tb]
\scriptsize
    \centering
    \begin{minipage}{\textwidth}
    \tablestyle{6pt}{1.1}
    \begin{tabular}{y{20mm}x{20mm}x{20mm}x{20mm}x{30mm}x{30mm}x{30mm}}
        \toprule
       & MonST3R  & DROID & SmallGS &  SmallGS w/ 3 DINOv2 & SmallGS w/ 16 DINOv2  \\
        \midrule
        $ATE\downarrow$   & 0.00294 & 0.00268 & 0.00263 &  \underline{0.00254} & \textbf{0.00228} \\
        $RPE_r\downarrow$ & \underline{0.316}   & 0.320   & \textbf{0.301}   &   0.335   & 0.340 \\
        $RPE_t\downarrow$ & 0.00186 & 0.00197 & 0.00183 &  \underline{0.00135} & \textbf{0.00123} \\
        $\Delta v\downarrow$ & 0.0258 & 0.0263 &0.0230 & \underline{0.00980} & \textbf{0.00090} \\
        \bottomrule
    \end{tabular}
    \end{minipage}
    \caption{
    Comparison of baseline methods and SmallGS based on $ATE$, $RPE_r$, and $RPE_t$. $\Delta v$ represents the average difference in camera translation velocity between the predicted and ground-truth trajectory. Bold values indicate the best performance, underlined values the second-best. DINOv2 denotes that SmallGS learns visual features instead of RGB in Gaussian Splatting, with 3 and 16 indicating the top 3 and 16 most informative feature channels, respectively.
    }
    \label{tab:gs_pose}
\end{table*}
In our SmallGS framework, we set the first frame in each video segment as the canonical frame and estimate the relative camera poses for the subsequent frames.
\subsection{Dataset}
We select 46 sequences from the TUM-dynamics dataset \cite{sturm2012benchmark}. These sequences feature very small camera baselines and include dynamic objects within the scenes.
The length of each sequence is 30-frame, and the image resolution is $512 \times 384$. The number of frames for each video segment in SmallGS is 15 for all experiments. We compare our estimated camera poses with the ground truth after alignment.

\subsection{Evaluation Metric}

We evaluate the camera pose estimation by reporting the Absolute Trajectory Error ($ATE$) and the Relative Pose Error ($RPE$) in both translation and rotation.

In addition, we compare the average difference in camera translation velocity ($\Delta v$) between our predicted trajectory and the ground truth. Velocity represents the rate of change in camera position. A high velocity difference between frames indicates significant camera drift and jitter in pose estimation. Conversely, similar inter-frame velocities between the ground-truth and estimated camera poses imply motion consistency.
\subsection{Baseline and ours}
Camera pose estimation is a well-established topic, with existing methods generally achieving strong performance. SmallGS is developed as an improvement over CF-3DGS   \cite{cf3dgs}, which estimates camera poses simultaneously with scene reconstruction as a baseline. Unlike CF-3DGS, SmallGS focuses exclusively on camera pose estimation and introduces a batched optimization approach. We compare our method against learning-based visual odometry approaches for dynamic scenes, specifically DROID-SLAM  \cite{teed2021droid} and MonST3R  \cite{monst3r}.

Our implementation is based on PyTorch and GSplat   \cite{gsplat}, following the default GSplat configuration unless otherwise stated. The Gaussian splatting of the canonical frame is removed after optimizing the camera pose for each sliding window. 
The depth and confidence masks predicted by MonST3R are used both for dense point cloud initialization in Gaussian splatting and as semantic masks during the learning process. We mask out the dynamic obejcts using the semantic masks in the Gaussian splatting. We initialize the camera pose using the identity matrix and use the same depth and confidence masks from MonST3R across all experiments to ensure fair comparisons.
While our experiments depend on MonST3R outputs for consistency, the SmallGS framework is not restricted to MonST3R specifically; alternative sources for depth, semantic masks, and intrinsics can be easily incorporated under different scenarios.


Furthermore, to extract more informative features than RGB images, we learn feature fields using Gaussian splatting   \cite{featuregs,gsplat}. We employ DINOv2   \cite{dinov2} to extract robust visual features, utilizing its distilled backbone in all subsequent experiments. Principal Component Analysis (PCA) is applied to filter the most informative channels from patched tokens, selecting the top sixteen features for comparison with the same model trained on RGB images. 
\section{Results and Discussion}
\subsection{Experiment Results}
The results are shown in Tab.\ref{tab:gs_pose}. SmallGS improved trajectory accuracy without additional features. Both the relative pose error ($RPE$) and absolute trajectory error ($ATE$) outperform MonST3R and DROID-SLAM, demonstrating that batched Gaussian splatting effectively learns camera trajectories. Additionally, the lower $\Delta v$ indicates that our estimated trajectory closely aligns with the ground truth motion. This is because the Gaussian splatting is consistent over small viewpoint changes, providing a robust batched camera pose optimization process which is minimally affected by noisy depth map or point cloud initialization.
The segment-wise camera pose estimation also reduces drift accumulation.  

Incorporating DINOv2 visual features further enhances accuracy. DINOv2 extracts general visual features that encode semantic and depth information. Using 3-channel DINOv2 feature maps significantly improves $ATE$ and $RPE_t$, highlighting the effectiveness of leveraging DINOv2 features. The experiment with 16-channel DINOv2 features achieves the best overall performance. Although $RPE_r$ is slightly higher than the baseline methods, improvements in $ATE$ and $RPE_t$ indicate that SmallGS provides high-precision camera position estimation. Furthermore, the lowest $\Delta v$ suggests that SmallGS with 16-channel DINOv2 features yields the most accurate camera motion alignment with the ground truth.  
Thus, we conclude that SmallGS effectively learns meaningful 3D visual features, enhancing scene representation and improving camera pose estimation.

Our method achieves lower $ATE$, $RPE$, and $\Delta v$ compared to MonST3R and DROID-SLAM overall, particularly improving the accuracy of relative translation in camera pose estimation. This suggests that SmallGS is better suited for small-baseline videos, benefiting from the robust scene geometry provided by Gaussian splatting in small-baseline video and the strong visual features extracted by DINOv2.

\begin{figure}[t]
    \centering
    \begin{subfigure}{0.49\linewidth}
        \centering
        \includegraphics[width=\linewidth]{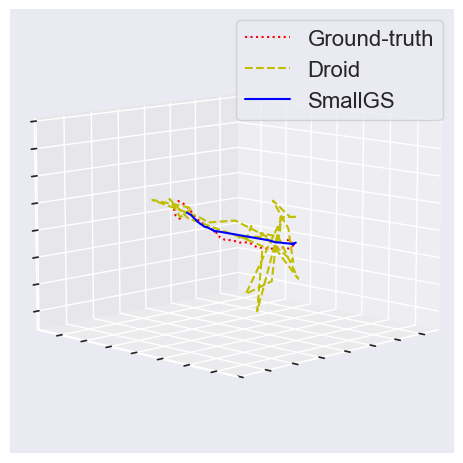}
        \caption{Comparison with DROID}
        \label{fig:exp_DROID}
    \end{subfigure}
    \hfill
    \begin{subfigure}{0.49\linewidth}
        \centering
        \includegraphics[width=\linewidth]{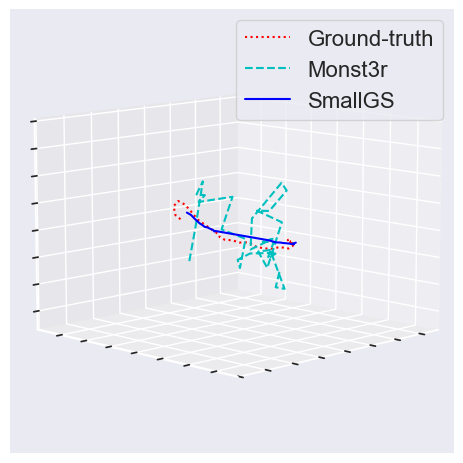}
        \caption{Comparison with MonST3R}
    \label{fig:exp_MonST3R}
    \end{subfigure}

    \caption{Fig.~\ref{fig:exp_MonST3R} compares the estimated camera trajectories of MonST3R and SmallGS with 16-channel DINOv2 features. Fig.~\ref{fig:exp_DROID} shows the same SmallGS trajectory compared to DROID. The red dashed line represents the ground truth.}
    \label{fig:experiment_comparison}

\end{figure}
\begin{figure*}
    \centering
    \includegraphics[width=0.9\linewidth]{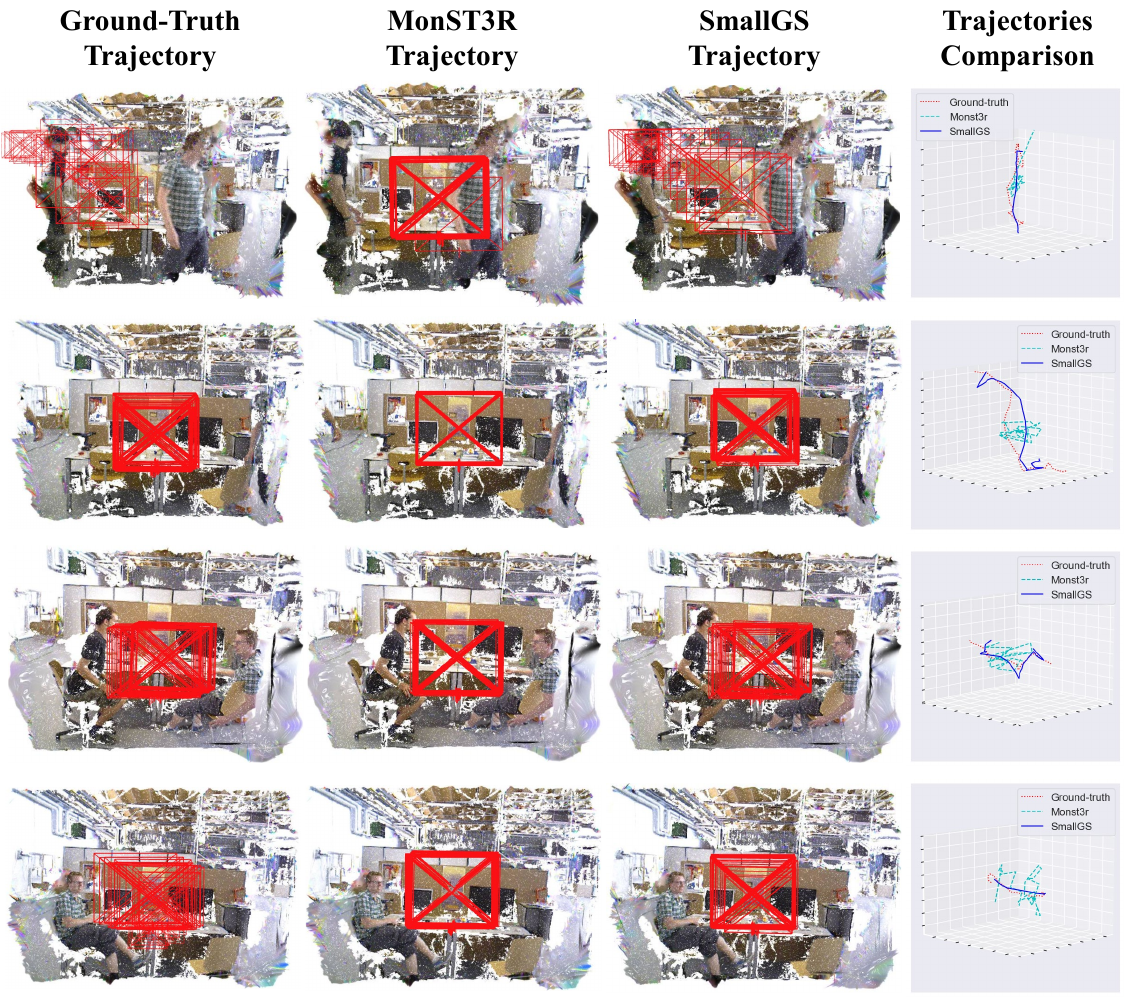}
    
    \caption{Comparison of ground-truth trajectories, MonST3R-predicted trajectories, and the SmallGS-learned trajectory with 16-channel DINOv2 feature maps. The trajectories predicted by MonST3R often exhibit jitter around the ground-truth trajectories. SmallGS efficiently learns the trajectories of small-baseline videos, improving camera pose accuracy.}
    \label{fig:demo}
    
\end{figure*}

Fig.~\ref{fig:experiment_comparison} illustrates the comparison among DROID-SLAM, MonST3R, and SmallGS trained with 16-channel DINOv2 features. Fig.~\ref{fig:demo} illustrates the comparisons of camera trajectories plotted in the 3D scenes, further illustrating that SmallGS produces smoother trajectores that have more consistent camera motion as the ground truth.  
The estimated camera poses from our method are smoother and exhibit less drift compared to the others.

\begin{table}[h]

    \centering
    \tablestyle{5pt}{1.1}
    \begin{tabular}{m{1.3cm}ccccc}
    \toprule
     & Time $\downarrow$ & $ATE$ $\downarrow$ & $RPE_r$ $\downarrow$ & $RPE_t$ $\downarrow$ & $\Delta v$ $\downarrow$ \\ 
    \midrule
    CF-3DGS & 682s & 0.00230 & \textbf{0.259} & 0.00189 & 0.0269 \\ 
    CF-3DGS +Mask & 635s & 0.00258 & 0.286 & 0.00182 & 0.0229 \\ 
    SmallGS & \textbf{271s} & \textbf{0.00228} & 0.340 & \textbf{0.00123} & \textbf{0.0009} \\ 
    \bottomrule
    \end{tabular}
    
    \caption{Average training time and accuracy comparison between CF-3DGS  \cite{cf3dgs} with and without semantic masks, and SmallGS w/ 16 DINOv2. }
    \label{tab:time}
    
\end{table}
In Tab.~\ref{tab:time}, we present the time consumption of SmallGS and CF-3DGS \cite{cf3dgs} for estimating the camera poses of 30-frame video sequences. CF-3DGS estimates pairwise camera poses by reconstructing the Gaussian splatting for the previous frame at each estimation, making it time-consuming. While CF-3DGS considers full scene reconstruction, reconstructing the 3D scene for each frame is unnecessary for camera pose estimation. We additionally applied the semantic masks, removing the dynamic object for CF-3DGS. However, $ATE$, $RPE_t$, and $\Delta v$ are larger for CF-3DGS with and without semantic masks than for SmallGS, demonstrating that SmallGS outperforms CF-3DGS in estimating camera poses for small-baseline videos in a more efficient way. 
\subsection{Ablation}
In this section, we present several ablation experiments to analyze the rationale behind our framework. Firstly, we use SmallGS to fine-tune the camera poses predicted by MonST3R, revealing that initializing camera poses with the identity matrix yields better results. Secondly, we utilize the pairwise pointmaps predicted by MonST3R to initialize Gaussian splatting. MonST3R's pointmaps provide a dense point cloud that is temporally consistent within paired images. These pointmaps encode both geometric and potential semantic information, making them more informative and robust than point clouds lifted from depth maps.



\paragraph{MonST3R Pointmaps Instead of DINOv2 Feature Maps}
MonST3R is trained to predict pairwise pointmaps that maintain temporal consistency between paired images within a sliding window. These pointmaps, which serve as dense point clouds, preserve scene geometry more effectively within the sliding window. They incorporate richer geometric and potential semantic information than depth maps. The correlation of each 3D Gaussian is ensured to match the corresponding 3D Gaussian in the paired frame, reducing the complexity of estimating transformations between two Gaussian splatting representations. Hence, MonST3R pointmaps could provide a more reliable initialization for Gaussian splatting.

To select appropriate image pairs for generating pointmaps, we choose the pairs that are the furthest apart within the video segments in SmallGS. We compare the results of initializing Gaussian splatting using MonST3R pointmaps and depth maps, as shown in Tab.~\ref{tab:abl:intermediate}. SmallGS using pointmaps achieves the lowest relative rotation error, as their temporal consistency provides better initialization for point clouds, improving the geometry of Gaussian splatting. However, SmallGS with DINOv2 features achieves the best overall performance with more accurate camera position estimation, demonstrating that robust general visual features are more effective for small-baseline camera pose estimation.

\begin{table}[tb]
    \centering
    \tablestyle{4pt}{1.1}
    \begin{tabular}{lcccc}
    \toprule
    & $ATE\downarrow$ & $RPE_r\downarrow$ & $RPE_t\downarrow$ & $\Delta v\downarrow$ \\ 
    \midrule
    SmallGS & 0.00263 & \underline{0.301} & 0.00183 & 0.0230 \\ 
    SmallGS w/ pointmaps & \underline{0.00251} & \textbf{0.293} & \underline{0.00181} & \underline{0.0228} \\ 
    SmallGS w/ DINO & \textbf{0.00228} & 0.340 & \textbf{0.00123} & \textbf{0.0009} \\ 
    \bottomrule
\end{tabular}
    
    \caption{Comparison of SmallGS initialized with depth maps vs. MonST3R pointmaps. SmallGS uses depth maps without DINOv2, SmallGS w/ 16 DINO incorporates 16-channel DINOv2 features, and SmallGS w/ pointmaps is initialized with MonST3R pointmaps.}
    \label{tab:abl:intermediate}
    
\end{table}


\paragraph{Camera Pose Refinement}
Tab.\ref{tab:abl:refine_pose} presents a comparison between using SmallGS to refine the camera poses predicted by MonST3R and learning camera poses directly without a camera pose prior. In the refinement setting, we initialize the camera poses for each frame using the predicted camera poses from MonST3R.

The relative pose error of using SmallGS for refinement is lower. However, the $ATE$ and $RPE_t$ of SmallGS refining MonST3R's camera poses are worse than the non-refinement setting. Additionally, the larger difference in camera position velocity ($\Delta v$) indicates that the refined camera poses lack smoothness. This is because MonST3R predicts jittery and non-smooth camera poses in small-baseline videos, making them more challenging for SmallGS to fine-tune.
As shown in Fig.~\ref{fig:experiment_comparison}, the predicted trajectory of MonST3R exhibits noticeable jitter. Thus, initializing camera poses with the identity matrix proves to be more effective for small-baseline videos.

        


\begin{table}
    \centering
    \tablestyle{4pt}{1.1}
        
    \begin{tabular}{lcccc}
    \toprule
    & $ATE\downarrow$ & $RPE_r\downarrow$ & $RPE_t\downarrow$ & $\Delta v\downarrow$ \\ 
    \midrule
    Refine w/ 16 DINO & 0.00229 & \textbf{0.321} & 0.00136 & 0.00106 \\ 
    SmallGS w/ 16 DINO & \textbf{0.00228} & 0.340 & \textbf{0.00123} & \textbf{0.00090} \\ 
    \bottomrule
\end{tabular}
    \caption{Comparison of fine-tuning MonST3R-predicted camera poses vs. SmallGS-only estimation. Refine w/ 16 DINO fine-tunes poses using SmallGS with 16-channel DINOv2, while SmallGS w/ 16 DINO runs SmallGS without prior poses.
    }
    \label{tab:abl:refine_pose}
\end{table}


\section{Conclusion and Future Work}
We present SmallGS, a new approach to estimating temporally consistent camera motions in small-baseline videos with dynamic objects. Our approach improves the camera pose accuracy by incorporating self-supervised robust visual features besides RGB images.  
The experiment results on small-baseline videos in TUM-dynamics~\cite{sturm2012benchmark} show the effectiveness of SmallGS, and its superior performances to state-of-the-art baselines.

Although using prior camera poses from MonST3R results in slightly less accurate estimation, the difference between SmallGS with and without priors remains small. 
Future work could further explore the impact of incorporating prior camera poses in SmallGS. 

\myparagraph{Acknowledgement.}
We sincerely thank Michael J. Black for fruitful discussions on the small-baseline video scenario.
{
    \small
    \bibliographystyle{ieeenat_fullname}
    \bibliography{main}
}


\end{document}